\documentclass{article}

\usepackage{arxiv}

\usepackage{bm}
\usepackage[utf8]{inputenc} % allow utf-8 input
\usepackage[T1]{fontenc}    % use 8-bit T1 fonts
\usepackage{hyperref}       % hyperlinks
\usepackage{url}            % simple URL typesetting
\usepackage{booktabs}       % professional-quality tables
\usepackage{amsfonts}       % blackboard math symbols
\usepackage{nicefrac}       % compact symbols for 1/2, etc.
\usepackage{microtype}      % microtypography
\usepackage{lipsum}
\usepackage{amsmath,amssymb}
\usepackage{graphicx}

\usepackage{xspace}
\newcommand*{\eg}{\textit{e.g.}\@\xspace}
\newcommand*{\ie}{\textit{i.e.}\@\xspace}

\title{Localized Interactive Instance Segmentation}

\author{
  Soumajit Majumder\\
  Institute of Computer Science II\\
  University of Bonn, Germany\\
  \And
  Angela Yao\\
  School of Computing\\ National University of Singapore\\
  }

\begin{document}
\maketitle

\begin{abstract}
In current interactive instance segmentation works, the user is granted a free hand when providing clicks to segment an object; clicks are allowed on background pixels and other object instances far from the target object. This form of interaction is highly inconsistent with the end goal of efficiently isolating objects of interest. In our work, we propose a clicking scheme wherein user interactions are restricted to the proximity of the object. In addition, we propose a novel transformation of the user-provided clicks to generate a weak localization prior on the object which is consistent with image structures such as edges, textures etc. We demonstrate the effectiveness of our proposed clicking scheme and localization strategy through detailed experimentation in which we raise state-of-the-art on several standard interactive segmentation benchmarks.
\end{abstract}

%% INTRODUCTION
\section{Introduction}
Interactive object selection or interactive instance segmentation allows users to select objects of interest down to a pixel level by providing inputs such as clicks, scribbles and bounding boxes. The selected results are useful for downstream applications such as image/video editing~\cite{vos-wild,lazysnap}, medical diagnosis~\cite{med2,deepigeos}, image annotation tools~\cite{fluid,benenson19} etc. GrabCut~\cite{grabcut} is a pioneering example of interactive segmentation; other notable methods include Random Walk~\cite{randomwalk} and GeoS~\cite{geos}. 

More recent methods~\cite{twostream,risnet,majumder19,dextr,ifcn} have approached the problem with deep learning architectures such as convolutional neural networks (CNNs). In deep interactive segmentation, the input consists of the RGB image as well as \emph{`guidance'} maps based on user-provided supervision.  Users give `\emph{positive}' clicks on the object of interest and `\emph{negative}' clicks on the background or other objects in the scene.  The guidance map helps the network to focus on the object instance to segment; in an iterative setting, it helps to correct errors from previous segmentations~\cite{vos-wild,risnet,itis,majumder19,ifcn}. Typically, such guidance maps are generated via fixed rules and are not visible to the end user; only the image and intermediate and end segmentation results are visible to the interacting user.

For deep interactive segmentation, research efforts have predominantly been limited to introducing new architectures~\cite{vos-wild,itis} and more sophisticated training procedures~\cite{risnet,itis}.  Yet minimizing user interaction and maintaining high quality segmentation requires a fine interplay between good specification of user interactions and careful leveraging of the provided inputs.  Previous methods have ignored these aspects by allowing users to freely provide inputs~\cite{vos-wild,itis,ifcn} in any order and at any location in the scene.

In addition, the guidance generated from the clicks are primitive and agnostic to structures present in the image~\cite{vos-wild,itis,dextr}. In fact, the number and type of clicks to give, as well as how to encode user clicks are open research questions with enormous impact on the performance of the interactive system.  For example,~\cite{majumder19} showed that with improved click encodings, a simple base segmentation network such as the FCN-8s~\cite{fcn} can outperform methods~\cite{vos-wild,itis,dextr} that use much deeper and stronger base networks such as ResNet-101~\cite{resnet}.  In this paper, we follow along this line of work in looking at how to cleverly specify and leverage user clicks to improve interactive instance segmentation. 

In this paper, our interest is in directing user clicks to weakly constrain the area of interest for interactive segmentation.  Limiting the spatial extent is advantageous both for the network and the user, \ie it tells the network which area to focus on for learning and also gives some indication of object scale; it also directs the user clicks to ambiguous locations which will most benefit from guidance. 

Directing user clicks to specify the location may seem like an obvious way for interaction but few works on interactive segmentation have done so to date. Instead, they favour hard constraints enacted by directly cropping out the bounding boxes derived from user-given inputs~\cite{dextr} or object detections~\cite{deepgc}.  This hard crop relies on highly specific user inputs such as extreme points~\cite{dextr} which may slow down the user interaction, or having pre-trained object detectors for the object classes of interest for segmentation~\cite{deepgc}.  We favour a simple approach, where we ask users to first roughly localize objects with the first two interactions, \eg on the two opposite corners in a bounding box, or clicking at the center of the object and one outside the object boundary.  We propose using these first two interactions or clicks as the initial form of interaction. Ensuing corrective positive and negative clicks are constrained to be outside and within the enclosing boundary. 

In addition, we propose a new transformation scheme for the user-provided clicks which provides a weak localization prior on the object of interest and is consistent with low-level structures such as edges, textures, etc in the scene. Unlike~\cite{deepgc}, this prior is generated without using class-specific bounding box detections. With the arrival of newer clicks, this proposed transformation gradually refines the localization prior. Our proposed approach can deal naturally with several types of guidance modalities, including superpixel-based guidances~\cite{majumder19} and bounding box type guidances~\cite{deepgc}. Our key contributions are:
\begin{itemize}
    \item a simple yet efficient clicking scheme which focuses the user's attention to the object of interest and its vicinity,
    \item a novel transformation of the user clicks which provides a weak localization prior on the object; with the arrival of new user clicks the generated guidance map gradually refines to the object boundary. 
    \item state-of-the-art performance on three interactive image segmentation benchmarks including the challenging MS COCO~\cite{mscoco}; like other competing state-of-the-art methods in literature, through simulation of user clicks, we significantly reduce the amount of user input required to generate accurate segmentation. 
\end{itemize}

%%%%%%%%%%%%%%%%%%% RELATED WORKS
\section{Related Works}
The development of automated semantic and instance segmentation frameworks is a rapidly growing area in computer vision~\cite{deeplabv3,discrimfeat,contextenc}.  Accompanying this line of work is \emph{interactive} segmentation - where users give clicks, scribbles, or bounding boxes to adjust and improve the outputs of these fully automated methods.

Early interactive image segmentation approaches include  parametric active contours, snakes~\cite{snakes} and intelligent scissors~\cite{scissors}.  Since these methods focus primarily on boundary properties, they suffer when edge evidence is weak.  More recent methods are based on graph cuts~\cite{graphcuts,lazysnap,grabcut,growcut}, geodesics~\cite{geodesicmatting,geos}, or a combination of  both~\cite{geodesic,geodesicgraphcut}. However, since these methods try to separate foreground and background solely based on low-level features such as colour and texture, they are not robust and fare poorly when segmenting images with similar foreground and background appearances, intricate textures, and poor lighting.

Recently, deep convolutional neural networks (CNNs) have been incorporated into interactive segmentation frameworks.  The initial work of~\cite{ifcn} uses Euclidean distance maps to represent user-provided positive and negative clicks which are then concatenated with the original colour image and provided as input to a fully convolutional network~\cite{fcn}. Following works have focused primarily on making extensions with newer CNN architectures~\cite{vos-wild,itis} and iterative training procedures~\cite{risnet,itis}. Instead of training with fixed user clicks as input~\cite{ifcn}, iterative training algorithms~\cite{risnet,itis} progressively add clicks based on the error of the network predictions. 

In the majority of interactive segmentation frameworks, user guidance has been provided in the form of point-wise clicks~\cite{twostream,latentdiversity,risnet,itis,dextr,ifcn} which are then transformed into a Euclidean distance map~\cite{twostream,latentdiversity,ifcn}.  One observation made in~\cite{vos-wild,itis,dextr} was that encoding the clicks as Gaussians led to performance improvement because it localizes the clicks better~\cite{itis} and can encode both positive and negative click in a single channel~\cite{vos-wild}. A more recent work~\cite{benenson19} observed encoding user clicks as small binary disks to be more effective than Gaussian and the Euclidean encoding. 

Different to~\cite{vos-wild,itis,ifcn,dextr,benenson19,risnet}, we use guidance maps which are consistent with the low-level image structures. Additionally, we propose a superpixel box guidance map which provides weak localization cues to the network. This is similar in spirit to~\cite{benenson19,dextr,deepgc} in which object bounding boxes are cropped out from extreme points specified by the user~\cite{dextr}, (loose) ground truth bounding boxes~\cite{benenson19} or object detections~\cite{deepgc}. Our work relaxes the hard constraint of~\cite{dextr}, wherein clicks have to be placed on the four extremities of the object and on the object boundary. Furthermore, unlike~\cite{deepgc}, our proposed superpixel box guidance is class-agnostic and does not require having pre-trained object detectors available. 

%%%%%%%%%%%%%%%%%%%%%%%%%%%% METHOD

\section{Proposed Method}
We adopt the common approach for interactive segmentation that has been used in previous deep learning-based frameworks~\cite{vos-wild,risnet,majumder19,itis,ifcn}. The user provides inputs on the original RGB image in the form of \emph{`positive'} and \emph{`negative'} clicks to indicate foreground and background respectively. The clicks are then encoded into guidance maps via transformations (Section~\ref{sec:distmap} and Section~\ref{sec:spmap}).

Typically, pixel values on the guidance map are a function of the pixel distance on the image grid to the points of interaction (see Fig.~\ref{fig:guidance}). This includes Euclidean~\cite{twostream,ifcn} and Gaussian guidance maps~\cite{vos-wild,itis,dextr}. However, such guidance maps are generated in an image-agnostic manner with the assumption that pixels in an image are independent of one another. Alternative variants take image structures such as superpixels~\cite{swipecut,tapnshoot,majumder19} and region-based object proposals~\cite{majumder19} into consideration for generating guidance maps. Guidance maps are then concatenated as additional channels to the input image and passed through the network~\cite{ifcn,risnet,dextr,tapnshoot,majumder19}. 

\subsection{Interaction Loop}
In previous works~\cite{vos-wild,twostream,itis,majumder19,ifcn}, the user has the liberty to provide clicks anywhere in the scene. This includes clicking on object instances far from the one of interest. Intuitively, a user interested in recovering an object instance from the scene would primarily fixate in the vicinity of the object of interest and focus more on delineating the object from the nearby background. Additionally, unconstrained clicks on the background and other objects fail to provide hints on the whereabouts of the object which calls for additional click sampling strategies are proposed~\cite{ifcn,vos-wild,itis} to ensure negative clicks encompassing the object. 

We propose a simple yet intuitive interaction framework. At the onset of interaction, the user provides a click at the center of the object of interest followed by another click on a background pixel in the vicinity of the object (see Fig.~\ref{fig:guidance}). This first pair of clicks is used to generate a coarse prior on the location of the object (see Sec.~\ref{sec:spmap}) in the form of an enclosing box. We then restrict the locations of user subsequent inputs.  More specifically, negative clicks need to be given inside the estimated bounding box, while positive clicks need to be given outside. In turn, the new positive clicks are then used to update the location prior. 

Let us denote the set of positive and negative clicks as $\{\bm{c}^+_i\}$ and $\{\bm{c}^-_i\}$ respectively for $i=\{1,\cdots,n\}$ and the initial foreground and background click as $\bm{c}^+_0$ and $\bm{c}^-_0$ respectively. Based on $\{\bm{c}^+_0, \bm{c}^-_0\}$, a coarse prior $\mathcal{G}$ on the object location is generated (see Sec.~\ref{sec:spmap}).  We then restrict the locations of user subsequent inputs.  More specifically, negative clicks need to be given inside the estimated object location, while positive clicks need to be given outside. The new positive clicks are used to update the bounding box boundaries $\tilde{e}_0$ and $\tilde{e}_1$, while all additional clicks, $\bm{c}^+_{i\neq 0}$ and $\bm{c}^-_{i\neq 0}$, are used to update $\mathcal{G}$. 

\subsection{Superpixel-based Guidance Maps}\label{sec:distmap}
Superpixels are known for their ability to group locally similar pixels~\cite{nlc,topdownbottomup,fst}. For our guidance maps, we consider the superpixel-based variant of~\cite{majumder19} which outperformed approaches using Euclidean and Gaussian guidance maps. In~\cite{majumder19}, user clicks given at single pixels are propagated to entire superpixels. Guidance values of other superpixels in the scene are then given by the minimum Euclidean distance from the centroid of each superpixel to the centroid of a user-selected superpixel. Example superpixel-based guidance maps are shown in Fig.~\ref{fig:guidance}.

More specifically, let $\{\mathcal{Z}\}$ denote the set of superpixels constituting an image and $f_{\mathcal{Z}}^p$ denote a function which maps every pixel $p$ to its corresponding superpixel. Let $\{z^+\}=f_{\mathcal{Z}}^p(\{\bm{c}^+\})$ and $\{z^-\}=f_{\mathcal{Z}}^p(\{\bm{c}^-\})$ be the set of positive and negative superpixels based on the user-provided clicks. The value of each pixel for the guidance map $\mathcal{S}^+_{\mathcal{Z}}(p)$ corresponding to the set of positive clicks $\{\bm{c}^+\}$ is given by, 
\begin{equation}\label{eq:spdist}
    \mathcal{S}^+_{\mathcal{Z}}(p) = \min_{z\in\{z^+\}} d_c^2(z, f_{\mathcal{Z}}^p(p)),
\end{equation}
and likewise for $\mathcal{S}^-_{\mathcal{Z}}(\cdot)$ for $\{\bm{c}^-\}$. In Equation~\ref{eq:spdist}, $d_c^2(z_i, z_j)$ is the Euclidean distance between the centroids $z_i^c$ and $z_j^c$ of superpixels $z_i$ and $z_j$ respectively, where $z_i^c=(\sum_i x_i/|z_i|$, $\sum_i y_i/|z_i|)$ and $|z_i|$ is the number of superpixels in $z_i$. The values of the guidance maps are truncated to $255$. Examples of such guidance maps are shown in Fig.~\ref{fig:guidance}. 

We additionally experimented with guidance maps generated based on the CIE-LAB color difference between the annotated superpixels and the other superpixels as per~\cite{tapnshoot} but we did not observe any promising results. 

\subsection{Superpixel-box Guidance Map}\label{sec:spmap}
Cropping images to exactly contain the object of interest has been shown to improve interactive segmentation performance~\cite{benenson19,dextr}. However, such frameworks are limited by the placement of the additional clicks; ~\cite{dextr} requires corrective clicks to be placed precisely on object boundaries. Besides, this also leads to a set of unnatural training images dominated by the object of interest. This prevents the network from learning from the background regions. Unlike~\cite{dextr}, we refrain from cropping the image to contain only the object of interest.

\begin{figure}[t!]
    \centering
    \includegraphics[height=13.8cm,width=10.2cm]{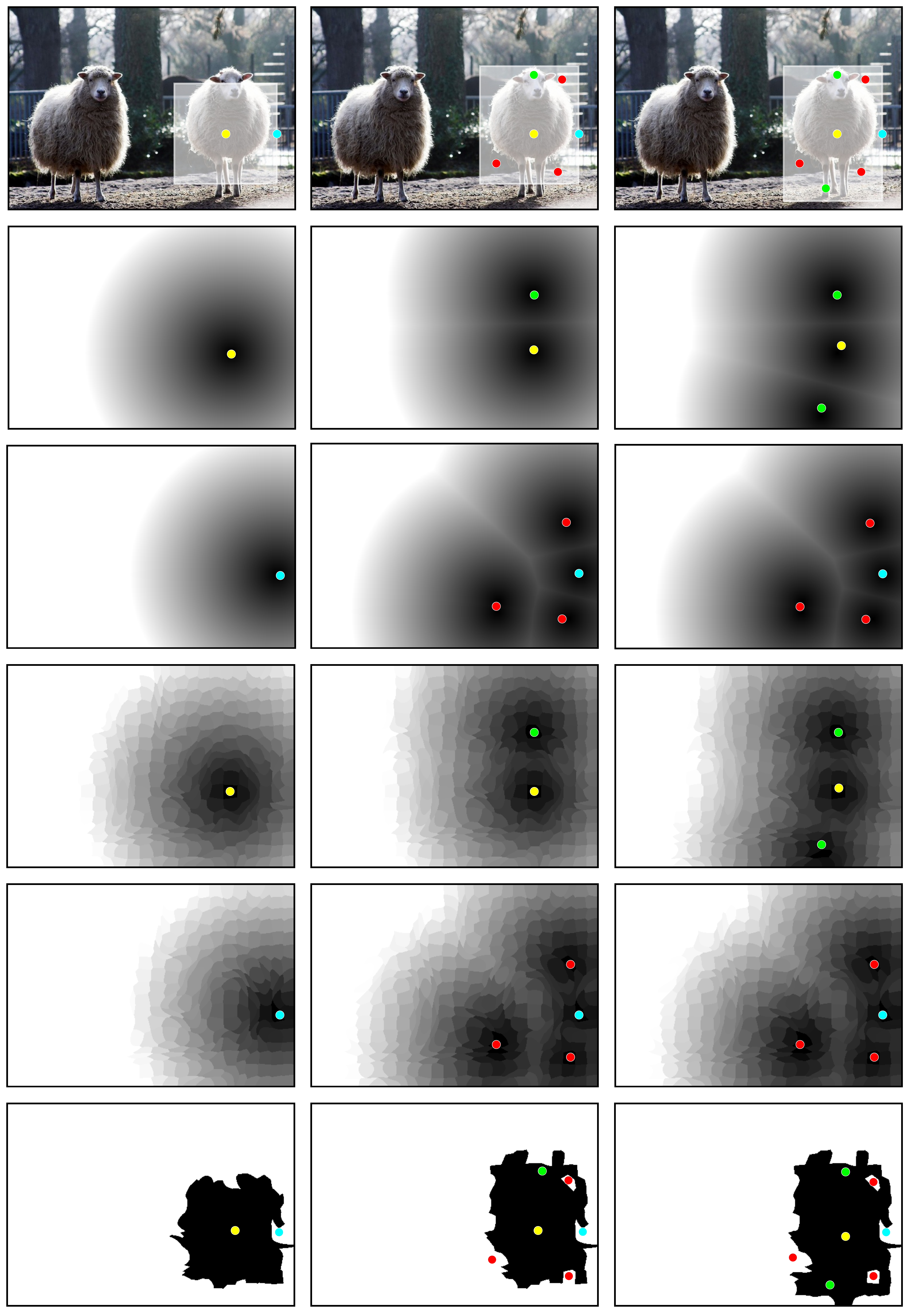}
    \caption{\textbf{Examples of Guidance maps}: At the onset of interaction, our approach receives an initial pair of clicks (denoted by yellow and blue at the center of the object and on a background pixel respectively). These clicks are transformed into guidance maps and used as input for the segmentation network. With each round of corrective clicking, the guidance maps are updated by considering the positive clicks (shown in green) and negative clicks (shown in red). Examples of user click transformations are shown in rows $2$ to $6$; rows $2$-$3$: positive and negative Euclidean distance maps, rows $4$-$5$: positive and negative superpixel-based guidance maps, row $6$: the superpixel box guidance. The values of the superpixel box guidance are inverted for ease of visualization. Note that euclidean distance maps are not used as guidance maps in our approach.}
    \label{fig:guidance}
\end{figure}

Instead, we provide, as an additional guidance channel, a weak prior on the whereabouts of the object in the scene based on the initial pair of clicks. At the onset, it behaves like a weak bounding box albeit consistent with low-level image features. With the arrival of additional clicks, it gets further refined into sloppy contours~\cite{sufficientannot}, and provides the segmentation network with a strong cue on the location of the objects (see Fig.~\ref{fig:guidance}). Unlike object-based guidance maps~\cite{majumder19} generated based on object proposals~\cite{mcg}, our proposed guidance is more flexible and adapts more quickly to the user-provided inputs.  

More formally, given the first pair of positive and negative click $\bm{c}^+_0=(x^+_0,y^+_0)$ and $\bm{c}^-_0=(x^-_0,y^-_0)$ for an image of size $w\times h$, we obtain the top-left and the bottom-right co-ordinates of the object, $e_0$ and $e_1$ respectively. Let $\{\mathcal{Z}_b\}\subset\{\mathcal{Z}\}$ be the set of superpixels which lie on or inside the spatial extent defined by $e_0$ and $e_1$. The value of each pixel $p$ of the superpixel-box based guidance map is given by,
\begin{equation}\label{eq:spbox}
    \mathcal{G}(p)= \bm{1}[p \subset z]\cdot\bm{1}[z \subset \{\mathcal{Z}_b\}]
\end{equation} 
where $\bm{1}[p \subset z]$ is an indicator function which returns $1$ if pixel $p$ lies belongs to superpixel $z$. $\bm{1}[z \subset \{\mathcal{Z}_b\}]$ returns $1$ if superpixel $z$ belongs to the set of superpixels $\{\mathcal{Z}_b\}$. For the additional set of clicks $\{\bm{c}^+_{i\neq 0}\}$ and $\{\bm{c}^-_{i\neq 0}\}$, we obtain the updated guidance map $\hat{\mathcal{G}(p)}$ as follows, 
\begin{align}
    \{z^+_{i\neq 0}\} &= f_{\mathcal{Z}}^p(\{\bm{c}^+_{i\neq 0}\}) \\
    \{z^-_{i\neq 0}\} &= f_{\mathcal{Z}}^p(\{\bm{c}^-_{i\neq 0}\}) \\
    \hat{\{\mathcal{Z}_b\}} &= \{\mathcal{Z}_b\}\cup \{z^+_{i\neq 0}\} \setminus \{z^-_{i\neq 0}\} \\
    \hat{\mathcal{G}(p)} &= \bm{1}[p \subset z]\cdot\bm{1}[z \subset \hat{\{\mathcal{Z}_b\}}]
\end{align}

\subsection{Simulating User Interactions}
To train and test our network, we simulate user interactions, as per previous works on interactive segmentation~\cite{vos-wild,itis,ifcn}. For simulating user interactions, we make use of the ground truth masks of PASCAL VOC $2012$~\cite{pascal} along with the additional masks from Semantic Boundaries Dataset (SBD) ~\cite{sbd}. We use the centroid of the ground truth masks as our first positive click; for concave object masks, clicks falling outside the object mask are relocated to a point within the object. We then displace the click location by $20$-$50$ pixels randomly; we ensure that the final click location remains within the object. This is done to introduce variation in the training data; the perturbation prevents center clicks to always fall on the same superpixel during each training iteration and also better approximates true user interactions which may not perfectly localize the object center.

Next, we sample the first negative click which is at least $d$ pixels away from the center click; in our experiments for a bounding box of height $h$ and width $w$, we set $d$ to be, 
\begin{equation}
    d=(r_1-r_2)\cdot w + (1+r_2)\cdot h
\end{equation}
where $r_1$ is sampled from the uniform distribution $\mathcal{U}(0,1)$ and $r_2$ is sampled from the normal distribution $\mathcal{N}(0,1)$. We use the first pair of clicks to generate the enclosing superpixel box; we keep superpixel boxes with an intersection over union (IoU) of $\geq0.7$. For simulating additional positive and negative clicks, we randomly pick $2$-$5$ superpixels from the set of superpixels lying outside the enclosing box and inside respectively.  

%%%%%%%%%%%%%%%%%%%%%%%%%%%%%%%%%%%%%%%%%%%%%%%% 
\section{Experimentation}
\subsection{Datasets and Evaluation}
We evaluate the performance our proposed method on five publicly available datasets used for benchmarking interactive image and video segmentation~\cite{tapnshoot,risnet,majumder19,dextr,ifcn}: PASCAL VOC 2012~\cite{pascal}, GrabCut~\cite{grabcut}, Berkeley~\cite{berkeley}, MS COCO~\cite{mscoco} and DAVIS-2016~\cite{davis2016}. \\ 

\noindent\textbf{PASCAL VOC 2012} consists of $1464$ training and $1449$ validation images across $20$ object classes; many images contain multiple objects. For training, we consider the $1464$ images plus the additional instance annotations from SBD provided by~\cite{sbd} which results in around $20,000$ instances across $20$ object classes.\\ 

\noindent \textbf{GrabCut} is a small dataset ($50$ images) and is one of the simpler interactive segmentation benchmark. The images consist of a single foreground object mostly with a distinctive appearance. \\

\noindent\textbf{Berkeley} consists of $100$ images with a single foreground object. The images in this dataset are representative of the typical challenges encountered in interactive segmentation such as heavily textured backgrounds, low contrast between the foreground object and the background, etc. \\
	
\noindent\textbf{MS COCO} is a large-scale image segmentation dataset with $80$ different object categories, $20$ of which are common with Pascal VOC $2012$. For fair comparison with~\cite{risnet,majumder19,ifcn}, we split the dataset into the $20$ PASCAL VOC $2012$ categories and the $60$ additional categories, and randomly sample $10$ images per category for evaluation.\\

\noindent\textbf{DAVIS-2016} is a dataset for video object segmentation. It consists of $50$ video sequences $20$ from which are in the validation set. The sequences feature a single foreground object; the pixel mask of the object is provided for all frames. \\

\noindent\textbf{Evaluation} The performance of fully automated instance segmentation algorithms is usually measured by the average mean intersection over union (mIoU) between the ground truth and the predicted object mask.  In interactive segmentation, a user can always provide more positive and negative clicks to further improve the predicted segmentation.  The established way of evaluating an interactive system is based on the number of clicks required for each object instance to achieve a fixed mIoU~\cite{vos-wild,itis,majumder19,ifcn}. This fixed mIoU threshold is 90\% for GrabCut and Berkeley and 85\% for the more challenging Pascal VOC $2012$ and MS COCO. Like~\cite{vos-wild,risnet,itis,majumder19,ifcn}, we threshold the maximum number of clicks per instance to $20$ clicks. Unlike~\cite{vos-wild,risnet,ifcn}, we do not apply any post-processing with a conditional random field.

\subsection{Implementation Details}

\textbf{Model Architecture} As our base segmentation network, we use DeepLab-v2~\cite{deeplabv2}; it consists of a ResNet-101~\cite{resnet} backbone and a Pyramid Scene Parsing network~\cite{psp} acting as the prediction head. The output of the CNN is a probability map representing whether a pixel belongs to the object. We initialize the weights from a network DeepLab-v2 model pre-trained on ImageNet~\cite{imagenet}, and fine-tuned on PASCAL VOC $2012$ for semantic segmentation. \\

\noindent\textbf{Training Data}
We further tune the network for instance segmentation on the $1464$ training images of PASCAL VOC $2012$~\cite{pascal} with the instance-level masks, along with the $10582$ images of SBD~\cite{sbd}. We further augment the training samples with random scaling, flipping and rotation operations. \\

\noindent\textbf{Superpixels}
We use SLIC~\cite{slic} as our superpixeling algorithm. We generate around $1000$ superpixels on an average per image; using $1000$ SLIC superpixels over $500$ SLIC superpixels have been shown to improve performance~\cite{majumder19}. Generating finer superpixels ($\geq 2000$ superpixels per image) degrades the performance as the superpixel-based guidance map degenerates to the Euclidean distance map. During evaluation on the GrabCut~\cite{grabcut}, Berkeley~\cite{berkeley}, MS COCO ~\cite{mscoco} and DAVIS-2016~\cite{davis2016} dataset, we roughly generate $1000$ SLIC superpixels~\cite{slic} for each image.\\

\noindent\textbf{Training Details}
Our network is trained to minimize a pixel-wise binary cross-entropy loss between the ground truth mask and the predicted mask.  For optimization, we use stochastic gradient descent with Nesterov momentum with its default value of $0.9$. The learning rate is fixed at $10^{-8}$ across all epochs and weight decay is $5\cdot10^{-4}$. A mini-batch of size $5$ is used. The implementation is done in PyTorch and built on top of the implementation provided by~\cite{dextr}. We train our network for $50$ epochs. \\

\noindent\textbf{Guidance Dropout} 
Dropout can be incorporated into the guidance inputs by introducing fixed-value maps into the training scheme with some probability.  Guidance dropout has been shown to be effective for interactive segmentation~\cite{majumder19} since it encourages the base segmentation network which is trained for semantic segmentation to switch over to instance segmentation without any user interaction. Following~\cite{majumder19}, during training, when the network receives an image with single object, we fix the value of $255$ for the superpixel-based and the superpixel-box based guidance map with a probability of $0.1$ to encourage good initial segmentations in absence of clicks. Additionally to make it robust to the number of user clicks, during training, we provide guidance maps with a single positive click (at the center) and the initial positive-negative click pair with a probability of $0.1$.  

\subsection{Ablation Studies}
We perform an ablation study to analyze the impact of different components in our interactive instance segmentation pipeline on the Berkeley dataset~\cite{berkeley} (see Table~\ref{tab:ablation}). Similar to the observation in~\cite{majumder19}, using a superpixel-based guidance map leads to a significant improvement over its euclidean distance map counterpart (denoted by EU) as used in iFCN~\cite{ifcn} (Table~\ref{tab:ablation}, rows 1-2). We observe additional gains from adopting the more recent ResNet-101~\cite{resnet} as our backbone architecture w.r.t FCN-8s~\cite{fcn} as used in~\cite{majumder19} (Table~\ref{tab:ablation}, row 3). 
\begin{table}[ht!]
    \centering
    \setlength{\tabcolsep}{4pt}
    \begin{tabular}{ll|cl|c|l|l}
        \toprule
         EU{\hskip 0.2in} & SP{\hskip 0.2in} & {\hskip 0.1in}BBox{\hskip 0.1in} & SPBox{\hskip 0.2in} & {\hskip 0.1in}DT{\hskip 0.1in} & Base{\hskip 0.6in}        & Berkeley\\
            &    &      &       &    & Network     & @$90\%$  \\\hline
        \checkmark   &    &      &      &  & FCN-8s~\cite{fcn} &  8.65~\cite{ifcn} \\
    & \checkmark & & & & FCN-8s~\cite{fcn} & 6.67~\cite{majumder19} \\ \hline 
    & \checkmark & & & & ResNet-101~\cite{resnet} & 6.32 \\ \hline
    & \checkmark & \checkmark & &\checkmark & ResNet-101~\cite{resnet} & 5.49 \\
    & \checkmark & \checkmark & & & ResNet-101~\cite{resnet} & 5.26 \\ 
    & \checkmark & & \checkmark & &  ResNet-101~\cite{resnet} & 5.18 \\ 
    \bottomrule
    \end{tabular}
    \vspace{0.5em}
    \caption{Ablation Study on the Berkeley Dataset~\cite{berkeley}}
    \label{tab:ablation}
\end{table}

Next, the benefits of having a weak localization prior on the object of interest as an additional mode of guidance Table~\ref{tab:ablation}, rows 4-6). BBox refers to the rectilinear box drawn between corner pixel locations $e_0$ and $e_1$ generated from user clicks $\{\bm{c}^+_i\}$ and $\{\bm{c}^-_i\}$.  SPBox refers to the superpixel-box guidance generated from $\{\bm{c}^+_i\}$ and $\{\bm{c}^-_i\}$ (Sec.~\ref{sec:spmap}, Equations $2$-$6$). Having a weak localization prior is shown to improve results across the board; the improvement is higher when using the SPBox guidance. Additionally, we consider the distance transform (DT) of the BBox as guidance but the average number of clicks increase from $5.26$ to $5.49$. Throughout our experiments, we use the superpixel-based guidances and the superpixel-box guidance as our guidance maps. 

\begin{table*}[t!]
	\centering
	\setlength{\tabcolsep}{5pt}
	\begin{tabular}{lccccc}
	\toprule
	Method         & GrabCut   & Berkeley & VOC12   & MS COCO     & MS COCO\\
	               & @$90\%$   & @$90\%$  & @$85\%$       & seen@$85\%$ & unseen@$85\%$ \\ 
	\midrule
	iFCN~\cite{ifcn}         & 6.04 & 8.65 & 6.88 & 8.31 & 7.82\\
	RIS-Net~\cite{risnet}    & 5.00 & 6.03 & 5.12 & 5.98 & 6.44\\
	ITIS~\cite{itis}         & 5.60 & - & 3.80 & - & - \\
	DEXTR~\cite{dextr}       & 4.00 & - & 4.00 & - & - \\
	VOS-Wild~\cite{vos-wild} & 3.80 & - & 5.60 & - & - \\
	FCTSFN~\cite{twostream}  & 3.76 & 6.49   & 4.58 & 9.62 & 9.62 \\
	IIS-LD~\cite{latentdiversity} & 4.79 & - & - & 12.45 & 12.45 \\
	MLG~\cite{majumder19}    & 3.58 & 5.60 & \textbf{3.62} & 5.40 & 6.10\\ 
	BRS~\cite{jang2019interactive} & 3.60 & \textbf{5.08} & - & - & - \\ \hline
	\textit{Ours} & \textbf{3.46} & 5.18 & 3.70 & \textbf{5.15} & \textbf{5.70} \\
	\bottomrule
	\end{tabular}
	\vspace{0.5em}
	\caption{The average number of clicks required to achieve a particular mIoU. The best results are indicated in \textbf{bold}.}
	\label{tab:all}
%\vspace{-2em}
\end{table*}

\subsection{Comparison to State of the Art}
\begin{figure}[t!h]
    \centering
    \includegraphics[height=5.5cm,width=12cm]{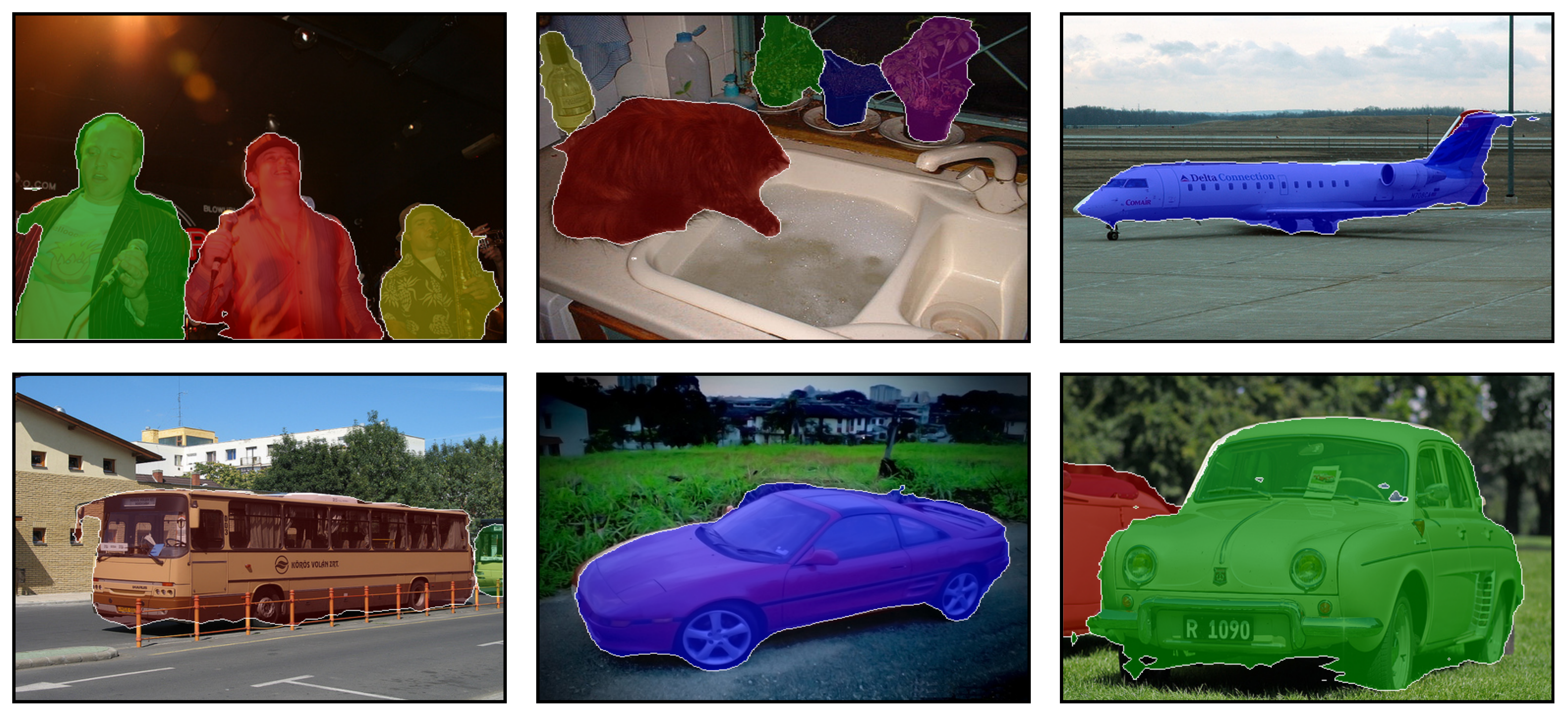}
    \caption{\textbf{Qualitative Results} Examples of high-quality object segmentations generated on PASCAL VOC $2012$. Note that final segmentation masks might not align to object boundaries as no CRF-based post-processing was performed.}
    \label{fig:qualitative}
\end{figure}

We compare the average number of clicks required to reach a required mIoU (see Table~\ref{tab:all}) against existing interactive segmentation approaches. We achieve the lowest number of clicks required for the GrabCut and for the challenging MS COCO (both seen and unseen categories) datasets, proving the benefits of restricting the interaction to only the object of interest. In Fig.~\ref{fig:qualitative}, we show some qualitative results from the PASCAL VOC $2012$ validation set.

As shown in Table~\ref{tab:all}, our full model needs the fewest number of clicks to reach the required mIoU threshold of $90\%$ on GrabCut, with a relative improvement of $3.3\%$. For MS COCO, we observe an improvement of $4.6\%$ and $6.5\%$ over the $20$ seen and $60$ unseen object categories respectively. We also report a relative $7.5\%$ improvement for PASCAL VOC $2012$ \textit{val} set w.r.t previous state-of-art algorithms using a fixed clicking scheme~\cite{dextr}. For MS COCO dataset, it should be noted that FCTSFN~\cite{twostream} and IIS-LD~\cite{latentdiversity} report their result averaged over all the $80$ object categories. 

\subsection{Correcting Masks for Video Object Segmentation}
Fully automated video object segmentation techniques can generate object segmentation masks of unsatisfactory quality; such masks are unsuitable for their intended downstream application. These scenarios can benefit from interactive segmentation approaches. Given an unsatisfactory prediction, users can provide additional clicks to improve the mask. Following~\cite{vos-wild,itis}, we proceed to improve the worst segmentation masks per sequence as generated by OSVOS~\cite{osvos}. The changed mIoU is reported after the addition of $1$, $4$ and $10$ clicks.

\begin{table*}[h]
    \setlength{\tabcolsep}{9pt}
	\centering
	\begin{tabular}{lcccc}
	\toprule
	Method & OSVOS\cite{osvos} & $1$-click  & $4$-clicks & $10$-clicks\\
	\midrule
	GrabCut\cite{grabcut} & $50.4$   & $46.6$ & $53.5$ & $68.8$\\
	iFCN\cite{ifcn}         & $50.4$ & $55.7$ & $71.3$ & $79.9$\\
	VOS-Wild\cite{vos-wild} & $50.4$ & $63.8$ & $75.7$ & $82.2$\\
	ITIS\cite{itis}         & $50.4$ & $67.0$ & $77.1$ & $82.8$\\\hline
	Ours                   & $50.4$  & \textbf{72.2} & \textbf{80.1} & \textbf{84.3} \\
  	\bottomrule
	\end{tabular}
	\vspace{0.5em}
	\caption{Refinement of the worst predictions from OSVOS~\cite{osvos} on DAVIS-2016~\cite{davis2016} (performance measured in mIoU).}
	\label{tab:vos}
%	\vspace{-2.5em}
\end{table*}

We initialize our enclosing area for the superpixel-box guidance map based on the initial segmentation by OSVOS. Superpixel-based guidance maps are set to a value of $255$. We then provide additional positive and negative clicks to improve the mask quality which are then used to update the superpixel-based guidance maps. Our proposed algorithm reports a significant gain of over $5\%$ in mIoU for a single click and also outperforms the reported results for $4$ and $10$ clicks (see Table~\ref{tab:vos}).

%%%%%%%%%%%%%%%%%%%%%%%%%%%%%%%%%%%%%%%%%%%%%%%% 

\section{Conclusion}
In this paper, we demonstrate that limiting the extent of user interaction to only the object of interest can significantly 
reduce the amount of user interaction required to obtain satisfactory segmentations. Additionally, via experiments, we demonstrate the benefits of having a weak localization prior generated in the form of superpixel box guidance. Our proposed algorithm primarily faced difficulties when trying to segment occluded instances. In such cases, the superpixel box guidance overlaps significantly, making it difficult for the network to segment both the instances properly. \\

\noindent\textbf{Acknowledgement}
Research in this paper was partly supported by the Singapore Ministry of Education Academic Research Fund Tier 1. We also gratefully acknowledge NVIDIA’s donation of a Titan X Pascal GPU.

\bibliographystyle{unsrt}  
\bibliography{references}

\end{document}